\documentclass[sigconf,screen,preprint]{acmart}

\usepackage{algorithm}
\usepackage{algorithmic}
\usepackage{amsfonts}
\usepackage{multirow}
\usepackage{makecell}
\usepackage{orcidlink}
\usepackage{array}
\usepackage{tabularx}
\usepackage{hyperref}
\usepackage{bm}
\usepackage{ulem}[normal]

\definecolor{cvprblue}{rgb}{0.21,0.49,0.74}
\hypersetup{pagebackref,breaklinks,colorlinks,allcolors=cvprblue}

\definecolor{blue}{rgb}{0.21,0.49,0.74}

\renewcommand\footnotetextcopyrightpermission[1]{} 
\AtBeginDocument{%
  }

\setcopyright{acmlicensed}
\copyrightyear{2025}
\acmYear{2025}
\acmDOI{XXXXXXX.XXXXXXX}

\acmConference[CIKM]{Make sure to enter the correct conference title from your rights confirmation email}{Nov 10-14, 2025}{Coex, Seoul, Korea}

\begin{document}

\title{Sequential Difference Maximization: Generating Adversarial Examples via Multi-Stage Optimization}

\author{Xinlei Liu\(^1\)\orcidlink{0000-0001-6586-4457}, Tao Hu\(^{1,2,*}\)\orcidlink{0000-0001-7641-5622}, Peng Yi\(^{1,2}\), Weitao Han\(^{1,2}\), Jichao Xie\(^1\), and Baolin Li\(^1\)}

\affiliation{%
 \institution{\(^1\) Information Engineering University, Zhengzhou, China \\
 \(^2\) Key Laboratory of Cyberspace Security, Ministry of Education of China, Zhengzhou, China \\
 \(^*\) Corresponding Author: {\tt hutaondsc@163.com}}
 \country{}
}

\renewcommand{\shortauthors}{Xinlei Liu et al.}

\begin{abstract}

Efficient adversarial attack methods are critical for assessing the robustness of computer vision models. In this paper, we reconstruct the optimization objective for generating adversarial examples as "maximizing the difference between the non-true labels' probability upper bound and the true label's probability," and propose a gradient-based attack method termed \textbf{Sequential Difference Maximization} (\textbf{SDM}). SDM establishes a three-layer optimization framework of "cycle-stage-step." The processes between cycles and between iterative steps are respectively identical, while optimization stages differ in terms of loss functions: in the initial stage, the negative probability of the true label is used as the loss function to compress the solution space; in subsequent stages, we introduce the \textbf{Directional Probability Difference Ratio} (\textbf{DPDR}) loss function to gradually increase the non-true labels' probability upper bound by compressing the irrelevant labels' probabilities. Experiments demonstrate that compared with previous SOTA methods, SDM not only exhibits stronger attack performance but also achieves higher attack cost-effectiveness. Additionally, SDM can be combined with adversarial training methods to enhance their defensive effects. The code is available at \url{https://github.com/X-L-Liu/SDM}\footnote{The manuscript has been accepted for presentation at CIKM 2025. The DOI assigned to the article is \url{https://doi.org/10.1145/3746252.3760870}.}.

\end{abstract}



\maketitle

\section{Introduction}
\label{Sec1}

Deep neural networks (DNNs) are widely applied to image information processing \cite{LeC15}, encompassing various computer vision tasks such as image classification \cite{HeDee16}, object detection \cite{CheRGB21}, and semantic segmentation \cite{Jin21}. However, recent research reveals the vulnerability of DNNs \cite{Neo14,Goo15,Cro20,Wan24}: taking image classification as an example, they are highly susceptible to adversarial examples—subtly perturbed inputs designed to mislead model predictions. Such examples pose significant risks to safety-critical applications such as medical diagnosis and facial recognition. Thus, developing effective adversarial attack methods has become a prerequisite for rigorously evaluating model robustness and advancing defense technologies.  

Gradient-based adversarial attacks represent the most stringent evaluation paradigm for DNNs \cite{Akh21}, including single-step Fast Gradient Sign Method (FGSM) \cite{Goo15}, as well as multi-step iterative methods such as Projected Gradient Descent (PGD) \cite{Mad18}, Auto-PGD (APGD) \cite{Cro20}, and AutoAttack (AA) \cite{Cro20}. The fundamental principle of these methods is to formulate an optimization problem for generating adversarial examples and construct corresponding loss functions. Subsequently, they perform gradient ascent in the gradient direction of input information to generate adversarial examples with high loss values. However, we have identified that the current settings of optimization objectives and loss functions are unreasonable, leading to the generation of "non-adversarial examples with high loss values," a phenomenon that is elaborated in Sec.~\ref{Sec3}.

We reconstruct the adversarial example generation objective as "maximizing the difference between the non-true labels' probability upper bound and the true label's probability." Compared to the prior "minimize the true label's logit", the more complex solution space of the current objective makes it more difficult to optimize. We thus propose Sequential Difference Maximization (SDM) based on the idea of sequential optimization. SDM decomposes the solution process into multiple stages: (1) The initial stage aims to minimize the true label's probability and constructs the loss function accordingly, thereby compressing the solution space for reduced complexity. (2) Subsequent stages introduce the Directional Probability Difference Ratio (DPDR) loss function to adjust the distribution of non-true labels' probabilities for improving their upper bound. Each stage contains multiple identical iterative steps, and takes the previous stage's optimal solution as the initial solution. All stages are optimized strictly in the set order, forming an optimization cycle. Thus, SDM establishes a three-layer optimization framework of "cycle-stage-step". Experiments show that SDM outperforms previous SOTA methods in attack performance and cost-effectiveness. We summarize the main contributions as follows:

\begin{itemize}
\item We propose a gradient-based adversarial attack method known as \textbf{Sequential Difference Maximization} (\textbf{SDM}), which generates adversarial examples by reconstructing the optimization objective and introducing distinct loss functions in multiple ordered optimization stages.
\item We identify the phenomenon of "non-adversarial examples with high loss values," which reveals the irrationality in the setting of optimization objectives and loss functions in some prior adversarial attack methods.
\item Experiments demonstrate that SDM has higher attack performance and cost-effectiveness, significantly outperforming previous SOTA methods. It can also be combined with adversarial training methods to enhance their defense.
\end{itemize}

\begin{table*}[t]
\caption{Logits and probabilities of two adversarial examples generated from the same clean example.}\label{Tab1}
  \centering
  \newcolumntype{C}{>{\centering}X}
  \newcolumntype{L}[1]{>{\centering\arraybackslash}p{#1}}
  \renewcommand{\arraystretch}{0.99}
  \scalebox{0.87}{
  \begin{tabularx}{1.12\linewidth}{C|L{0.94cm}|*{10}{@{\hspace{0.1em}}L{1.2cm}@{\hspace{0.1em}}}|L{0.785cm}|L{1.37cm}|L{1.37cm}}
    \toprule
    \multirow{2}{*}{\makebox[0pt][c]{Input}} & \multirow{2}{*}{\makebox[0pt][c]{Output}} & \multicolumn{10}{c|}{\(k\)} & \multirow{2}{*}{\makebox[0pt][c]{Loss}} & \multirow{2}{*}{\makebox[0pt][c]{\scalebox{0.95}[1]{Pred. Label}}} & \multirow{2}{*}{\makebox[0pt][c]{\scalebox{0.95}[1]{Att. Result}}}\\
    & & 1 & 2 & 3 & \textbf{4} (\(y\)) & 5 & 6 & 7 & 8 & 9 & 10 & &\\
    \midrule
    \multirow{2}{*}{\makebox[0pt][c]{\(\boldsymbol{x}'_{(1)}\)}} & \makebox[0pt][c]{\(S_k\)} & 0.314 & -1.267 & -0.126 & \textbf{1.438} & 0.264 & 1.036 & 0.191 & -0.118 & -0.498 & -1.041 & \multirow{2}{*}{1.196} & \multirow{2}{*}{4} & \multirow{2}{*}{Failed}\\
    & \makebox[0pt][c]{\(P_k\)} & 9.83\% & 2.02\% & 6.33\% & \textbf{30.25\%} & 9.35\% & 20.24\% & 8.69\% & 6.38\% & 4.36\% & 2.54\% & & &\\
    \midrule
    \multirow{2}{*}{\makebox[0pt][c]{\(\boldsymbol{x}'_{(2)}\)}} & \makebox[0pt][c]{\(S_k\)} & -0.674 & -1.434 & -0.398 & 2.864 & -0.488 & \textbf{3.367} & -0.371 & -0.613 & -1.421 & -0.833 & \multirow{2}{*}{1.057} & \multirow{2}{*}{6} & \multirow{2}{*}{Successful} \\
    & \makebox[0pt][c]{\(P_k\)} & 1.01\% &  0.47\% &  1.33\% & 34.74\% & 1.22\% & \textbf{57.45\%} & 1.37\% & 1.07\% & 0.48\% & 0.86\% & & &\\
    \bottomrule
  \end{tabularx}
  }
\end{table*}

\section{Motivation}
\label{Sec2}

Let \(f\) be a multiclass classifier mapping inputs from \(\mathbb{R}^d\) to output scores in \(\mathbb{R}^K\) (\(K>2\)). \(d\) and \(K\) represent the dimension of the input space and the total number of labels, respectively. For a clean input example \(\boldsymbol{x}\in[0, 1]^d \subset\mathbb{R}^d\), the model produces a vector \(f(\boldsymbol{x})\in\mathbb{R}^K\) as the logit. Common adversarial example \(\boldsymbol{x}'\) can be expressed as
\begin{equation}
\operatorname*{\arg\max}_{k\in\{1,\dots,K\}} f_k(\boldsymbol{x}') \neq y \quad \text{s.t.} \quad \left\|\boldsymbol{x}'-\boldsymbol{x}\right\|_p \leq \epsilon,
\label{Equ1}
\end{equation}
\noindent where \(y\) and \(\epsilon\) represent the true label of the input example and the perturbation budget, respectively. \(\left\|\cdot\right\|_p\) denotes the \(\bm{\ell}_p\)-norm.

Eq.~\ref{Equ1} is a highly non-convex optimization problem that is difficult to solve directly \cite{Chr14}. Therefore, \textit{Goodfellow} et al. formulated the objective function to minimize the true label's logit \cite{Goo15}:
\begin{equation}
\min_{\boldsymbol{x}'} f_y(\boldsymbol{x}') \quad \text{s.t.} \quad \left\|\boldsymbol{x}' - \boldsymbol{x}\right\|_p \leq \epsilon.
\label{Equ2}
\end{equation}

Based on the aforementioned optimization objective, attack methods such as the Fast Gradient Sign Method (FGSM) \cite{Goo15}, Projected Gradient Descent (PGD) \cite{Mad18}, and Basic Iterative Method (BIM) \cite{Kur17} all reduce the confidence of adversarial examples in true labels by maximizing the value of the cross-entropy (CE) loss function.

\textbf{However, is it truly reasonable to set the objective function as "minimizing the logit or probability of the true label"?}

We argue that this setup is \textbf{unreasonable}. Next, we will elaborate on our viewpoint by discussing an interesting phenomenon, referred to as "non-adversarial examples with high loss values."

To simplify notation, we denote the logit \(f_k(\boldsymbol{x}')\) for label \(k\) as \(S_k\), and the normalized probability for label \(k\) as \(P_k\), expressed as:
\begin{equation}
P_k=\frac{e^{S_k}}{\sum_{i=1}^K{e^{S_i}}}.
\label{Equ3}
\end{equation}

As shown in Tab.~\ref{Tab1}, \(\boldsymbol{x}'_{(1)}\) and \(\boldsymbol{x}'_{(2)}\) are two adversarial examples generated from a clean example on CIFAR-10. The perturbation budget is \(\boldsymbol{\ell}_\infty=8/255\). The true label of the clean example is "\(\bm{4}\)". The loss is computed via PyTorch's CrossEntropyLoss function.

For example \(\boldsymbol{x}'_{(1)}\), the predicted label is the same as the true label, indicating a failed attack. Its true label's probability and loss are 30.25\% and 1.196, respectively. In contrast, example \(\boldsymbol{x}'_{(2)}\) has a predicted label different from the true label, representing a successful attack, with its true label's probability and loss being 34.74\% and 1.057, respectively. Example \(\boldsymbol{x}'_{(1)}\) corresponds to the non-adversarial example with high loss value mentioned previously.

This phenomenon indicates that low true label probability and high loss value are not sufficient conditions for generating adversarial examples. The optimization objective and loss function for generating adversarial examples need to be re-explored.

\section{Methodology}
\label{Sec3}

In this section, we reconstruct the optimization objective and propose a novel gradient-based attack method to encompass a broader generation of adversarial examples.

\subsection{Reconstruction of Optimization Objective}

Compared to the implicit constraint (\(P_y<P_k\), \(k\neq y\)) in Eq.~\ref{Equ2}, the constraint that the true label's probability is less than the non-true labels' probability upper bound (\(P_y<P_{\tau}\), \(\tau=\operatorname*{\arg\max}{\{P_k|k\neq y\}}\)) is more relaxed. Therefore, we reconstruct the optimization objective for generating adversarial examples as "maximizing the difference between the non-true labels' probability upper bound and the true label's probability," denoted as
\begin{equation}
\max_{\boldsymbol{x}'}\big\{P_{\tau}-P_y\big\} \quad \text{s.t.} \quad \left\|\boldsymbol{x}' - \boldsymbol{x}\right\|_p \leq \epsilon.
\label{Equ4}
\end{equation}

For the above optimization problem, some intuitive approaches include directly transforming the optimization objective into a loss function \cite{Car17}, or employing the Difference of Logits Ratio (DLR) loss \cite{Cro20}. Nevertheless, owing to the presence of two distinct optimization directions and the high non-convexity of the loss function, these methods are prone to being trapped in local optima.

\subsection{Sequential Difference Maximization}

Inspired by sequential optimization \cite{Jam92}, we propose the Sequential Difference Maximization (\textbf{SDM}) to solve the optimization problem in Eq.~\ref{Equ4}. SDM is a multi-stage iterative algorithm that sequentially optimizes the sub-objectives. Each stage utilizes the optimal solution from the previous stage as its initial solution, gradually approaching the optimal solution of the overall objective.

\begin{figure}[t]
  \centering
  \includegraphics[width=0.96\linewidth]{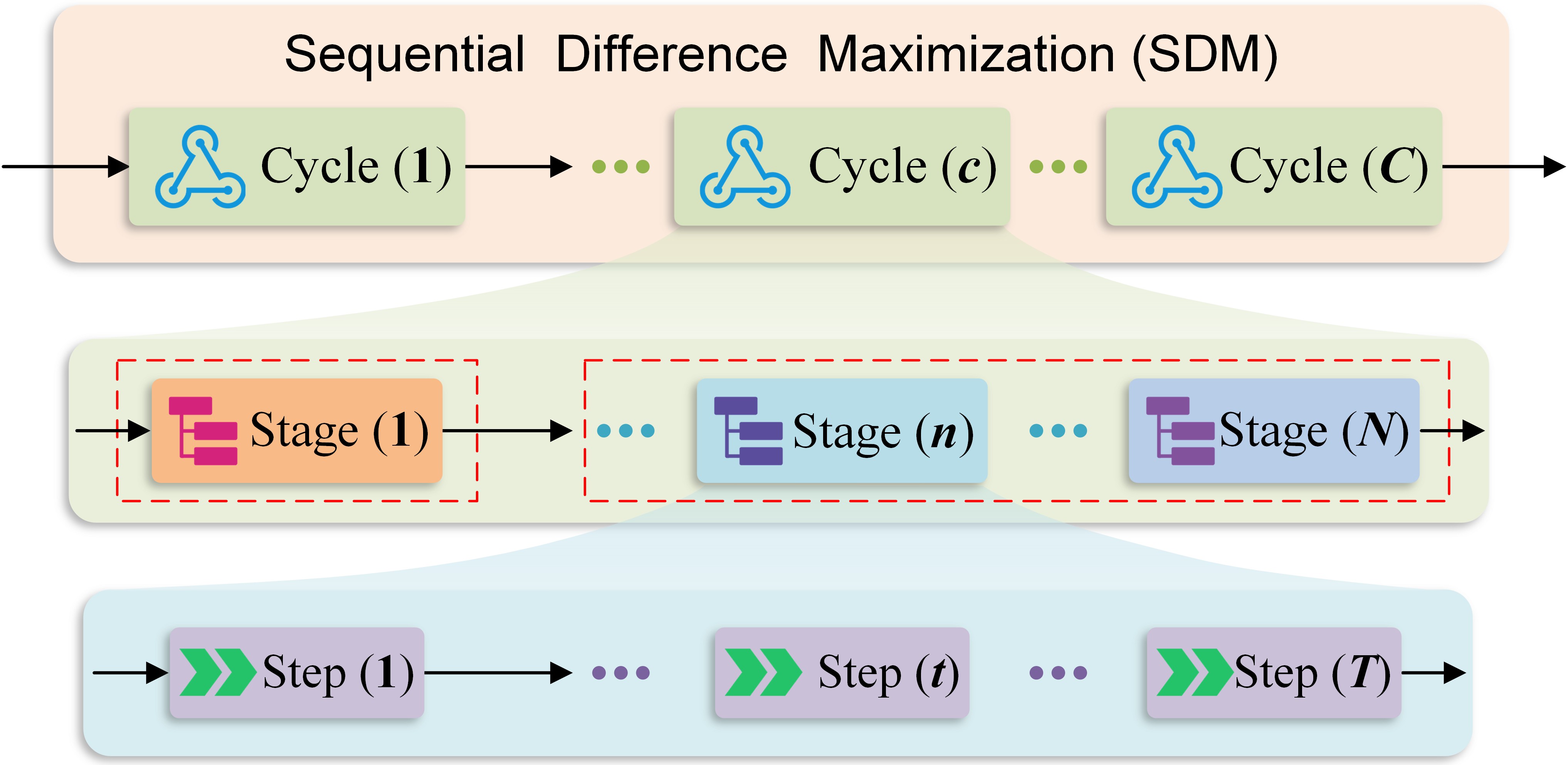}
  \caption{Overall structure of SDM.}
  \vspace{-5pt}
  \label{Fig1}
\end{figure}

\textbf{Overall structure.}  As shown in Fig.~\ref{Fig1}, the proposed SDM consists of \(C\) \dashuline{identical} optimization \textbf{cycles}, each of which contains \(N\) \uline{different} optimization \textbf{stages}. Each optimization stage consists of \(T\) \dashuline{identical} iteration \textbf{steps}. In the iteration steps, gradient ascent is performed to maximize the value of the loss function, thereby approaching the specific optimization sub-objective. 

The distinction between optimization stages lies in the fact that each stage has its unique sub-objectives and loss functions. SDM executes these optimization stages in the predetermined order.

\textbf{Sub-objective and loss function for initial stage.} We set the sub-objective for the initial stage (\(n=1\)) as "reducing the true label's probability." Accordingly, the negative probability loss is
\begin{equation}
L_y = -P_y.
\label{Equ5}
\end{equation}

Compared to optimizing \( P_\tau \), optimizing \( P_y \) first can reduce the solution space and facilitate subsequent optimization.

\textbf{Sub-objectives and loss functions for subsequent stages.} We set the sub-objectives for the \( n \)-th (\(2\leq n\leq N\)) stage as: (1) maintaining the true label's low probability (\(P_y\)); (2) reducing the \( n \)-th largest label's probability (\(\grave{P}_n\)) to increase the non-true labels' probability upper bound (\(P_{\tau}\)). To achieve this, we propose a loss function termed \textbf{Directional Probability Difference Ratio} (\textbf{DPDR}):
\begin{equation}
L_\text{DPDR}^{(n)}=\frac{P_\tau-P_y}{\delta-\operatorname{sign}(P_\tau-P_y)\cdot(P_\tau-\grave{P}_n-\delta)+\zeta}.
\label{Equ6}
\end{equation}

\begin{algorithm}[t]
	\caption{Detailed algorithm for SDM.}
	\label{Alg1}
	\begin{algorithmic}[1]
            \REQUIRE A multi-class classifier \(f\) with the parameter \(\theta\), a clean example \(\boldsymbol{x}\) with the true label \(y\), a adversarial perturbation budget \(\epsilon\) and step size  \(\alpha\) under the \(\boldsymbol{\ell}_p\)-norm constraint, the number of optimization cycles \( C \), the number of optimization stages \( N \), and the number of iteration steps \( T \).
		\ENSURE The adversarial example \(\boldsymbol{x}'\).
            \STATE Set the initial value of the adversarial example as \(\boldsymbol{x}'_0=\boldsymbol{x}\);
            \FOR{\(c=1\) to \(C\)}
                \FOR{\(n=1\) to \(N\)}
                    \FOR{\(t=1\) to \(T\)}
                        \IF{\(n\) \text{is} \(1\)}
                            \STATE Set the loss function as \(L=L_y\);
                        \ELSE
                            \STATE Set the loss function as \(L=L_\text{DPDR}^{(n)}\);
                        \ENDIF
                        \STATE Calculate the gradient of the adversarial example \(\boldsymbol{x}'_{t-1}\): \(Grad=\nabla_{\boldsymbol{x}'_{t-1}}L(\theta,\boldsymbol{x}'_{t-1},y)\);
                        \STATE Update the adversarial example and clip it to the specified perturbation boundaries: \\
                        \(\boldsymbol{x}'_t=\boldsymbol{x}+\text{clamp}(\boldsymbol{x}'_{t-1}-\boldsymbol{x}+\alpha\cdot\text{sgn}(Grad),-\epsilon,\epsilon)\);
                    \ENDFOR
                    \STATE Set the initial value of the adversarial example as \(\boldsymbol{x}'_0=\boldsymbol{x}'_T\).
                \ENDFOR
            \ENDFOR
	\end{algorithmic}
\end{algorithm}

In Eq.~\ref{Equ6}, \(\delta=0.5\times\max(P_\tau-\grave{P}_n)\) serves as a bias term to ensure that the denominator of the \(L_\text{DPDR}^{(n)}\) remains greater than zero. The parameter \( \zeta \) is set to a default value of \(10^{-10}\) to enhance the numerical stability. \(\grave{P}\) represents the descending order of \(P\), denoted as \(\grave{P}=\{\grave{P}_k\}\), satisfying \(\grave{P}_1\geq\grave{P}_2\geq\ldots\geq\grave{P}_K>0\).

When the attack is failed (\( P_\tau < P_y \)), the denominator of \(L_\text{DPDR}^{(n)}\) is \( P_\tau - \grave{P}_n+\zeta\). Increasing \(L_\text{DPDR}^{(n)}\) will raise the value of \(P_\tau\) while decreasing the value of \(\grave{P}_n\). Conversely, when the attack is successful (\(P_\tau>P_y\)), the denominator becomes \(2\delta-(P_\tau-\grave{P}_n)+\zeta\). 
Increasing \(L_\text{DPDR}^{(n)}\) will continue to elevate \(P_\tau\) and decrease \(\grave{P}_n\). The sign function preserves the consistency of DPDR's optimization direction, thereby mitigating oscillations during the optimization process.

\textbf{Workflow.} In the initial stage (Stage 1), SDM uses the negative probability loss \(L_y\) as the loss function. It performs gradient ascent through multiple iteration steps to search for optimal solution. This optimal solution will serve as the initial solution for the subsequent stages. In the subsequent stages, Stage 2 employs \(L_\text{DPDR}^{(2)}\) as the loss function and continues the iteration steps similar to that in Stage 1. Next, Stage 3 will use the optimal solution from Stage 2 as the initial solution, employs \(L_\text{DPDR}^{(3)}\) as the loss function, and continues the iteration steps outlined above. From Stage 2 to Stage \(N\), SDM gradually reduces the 2nd, 3rd, ..., up to \(N\)-th largest label's probabilities via the DPDR loss function, thereby increasing the non-true labels' probability upper bound. Each stage takes the optimal solution from the previous stage as its initial solution. All optimization stages are executed sequentially to form an optimization cycle. After completing one cycle, a new cycle is initiated to mitigate systematic blind spots in the optimization process. The detailed algorithm for SDM is presented in Algorithm~\ref{Alg1}.

\section{Experiments}
\label{Sec4}

\subsection{Experimental Setup}
\label{Subsec41}

This study focuses on generating untargeted adversarial examples in white-box scenarios. WideResNet-28-10 \cite{Zag16} and CIFAR \cite{Kri09} are used as the backbone model and evaluation dataset, respectively.

Defense methods under attack include Adversarial Training (AT) \cite{Mad18}, TRADES \cite{ZhaThe19}, MART \cite{Wan20}, HAT \cite{Rad22}, and LOAT \cite{Yin24}. All utilize PGD \cite{Mad18} with a perturbation budget of \(\boldsymbol{\ell}_\infty=8/255\), step size of \(\boldsymbol{\ell}_\infty=2/255\), and 10 iterations to generate adversarial examples for training. Detailed training parameters are provided in the code.

Attack methods competing with the proposed SDM include PGD \cite{Mad18}, C\&W \cite{Car17}, APGD-CE (APGD\(_1\)) \cite{Cro20}, APGD-DLR (APGD\(_2\)) \cite{Cro20}, and AA \cite{Cro20}. For APGD\(_1\) and APGD\(_2\), the restart count is set to 5, with the total iteration step calculated as the product of restart count and iteration steps. The correspondence between SDM's total steps and cycles, stages, and iteration steps is shown in Tab.~\ref{Tab2}. The total steps for PGD and C\&W are consistent with their iteration steps. AA uses the default settings in the original paper. Other attack configurations are described in respective subsections.

\begin{table}[t]
\caption{Detailed interpretation of total steps in SDM.}\label{Tab2}
  \centering
  \newcolumntype{C}{>{\centering}X}
  \newcolumntype{L}[1]{>{\centering\arraybackslash}p{#1}}
  \scalebox{0.87}{
  \begin{tabularx}{1\linewidth}{C|*{7}{L{0.59cm}}}
    \toprule
    \makebox[0pt][c]{Total Steps} & \makebox[0pt][c]{10} & \makebox[0pt][c]{20} & \makebox[0pt][c]{50} & \makebox[0pt][c]{100} & \makebox[0pt][c]{200} & \makebox[0pt][c]{500} & \makebox[0pt][c]{1000}\\
    \midrule
    \makebox[0pt][c]{Cycles} & \makebox[0pt][c]{1} & \makebox[0pt][c]{1} & \makebox[0pt][c]{2} & \makebox[0pt][c]{2} & \makebox[0pt][c]{4} & \makebox[0pt][c]{4} & \makebox[0pt][c]{5}\\
    \makebox[0pt][c]{Stages} & \makebox[0pt][c]{5} & \makebox[0pt][c]{5} & \makebox[0pt][c]{5} & \makebox[0pt][c]{5} & \makebox[0pt][c]{5} & \makebox[0pt][c]{5} & \makebox[0pt][c]{5}\\
    \makebox[0pt][c]{Steps} & \makebox[0pt][c]{2} & \makebox[0pt][c]{4} & \makebox[0pt][c]{5} & \makebox[0pt][c]{10} & \makebox[0pt][c]{10} & \makebox[0pt][c]{25} & \makebox[0pt][c]{40}\\
    \bottomrule
  \end{tabularx}
  }
\end{table}

\begin{table*}[t]
\caption{Error rate comparison (\%) of different defenses under various adversarial attack methods. Evaluation datasets are CIFAR-10 and CIFAR-100. Within the same category, the most effective attack result is boldfaced.}\label{Tab3}
  \centering
  \newcolumntype{C}{>{\centering}X}
  \newcolumntype{L}[1]{>{\centering\arraybackslash}p{#1}}
  \newlength{\colNarrow}  
  \setlength{\colNarrow}{0.56cm}
  \newlength{\colWide}    
  \setlength{\colWide}{0.56cm}
  \scalebox{0.87}{
  \begin{tabularx}{1.13\linewidth}{C|*{2}{L{\colNarrow}}*{2}{L{\colWide}}*{1}{L{\colNarrow}}|*{2}{L{\colNarrow}}*{2}{L{\colWide}}*{1}{L{\colNarrow}}|*{2}{L{\colNarrow}}*{2}{L{\colWide}}*{1}{L{\colNarrow}}|*{2}{L{\colNarrow}}*{2}{L{\colWide}}*{1}{L{\colNarrow}}}
    \toprule
    \multirow{3}{*}{\makebox[0pt][c]{Defense}} & \multicolumn{10}{c|}{CIFAR-10} & \multicolumn{10}{c}{CIFAR-100}\\
    & \multicolumn{5}{c|}{\(\boldsymbol{\ell}_\infty=8/255\)} & \multicolumn{5}{c|}{\(\boldsymbol{\ell}_2=1.0\)} & \multicolumn{5}{c|}{\(\boldsymbol{\ell}_\infty=8/255\)} & \multicolumn{5}{c}{\(\boldsymbol{\ell}_2=1.0\)}\\    
    & \makebox[0pt][c]{PGD} & \makebox[0pt][c]{C\&W} & \makebox[0pt][c]{\scalebox{0.9}[1]{APGD\(_1\)}} & \makebox[0pt][c]{\scalebox{0.9}[1]{APGD\(_2\)}} & \makebox[0pt][c]{SDM} & \makebox[0pt][c]{PGD} & \makebox[0pt][c]{C\&W} & \makebox[0pt][c]{\scalebox{0.9}[1]{APGD\(_1\)}} & \makebox[0pt][c]{\scalebox{0.9}[1]{APGD\(_2\)}} & \makebox[0pt][c]{SDM} & \makebox[0pt][c]{PGD} & \makebox[0pt][c]{C\&W} & \makebox[0pt][c]{\scalebox{0.9}[1]{APGD\(_1\)}} & \makebox[0pt][c]{\scalebox{0.9}[1]{APGD\(_2\)}} & \makebox[0pt][c]{SDM} & \makebox[0pt][c]{PGD} & \makebox[0pt][c]{C\&W} & \makebox[0pt][c]{\scalebox{0.9}[1]{APGD\(_1\)}} & \makebox[0pt][c]{\scalebox{0.9}[1]{APGD\(_2\)}} & \makebox[0pt][c]{SDM}\\
    \midrule
    \makebox[0pt][c]{AT\(^\text{\cite{Mad18}}\)} & \makebox[0pt][c]{54.23} & \makebox[0pt][c]{53.92} & \makebox[0pt][c]{54.94} & \makebox[0pt][c]{54.06} & \makebox[0pt][c]{\textbf{55.89}} & \makebox[0pt][c]{69.40} & \makebox[0pt][c]{68.29} & \makebox[0pt][c]{69.20} & \makebox[0pt][c]{67.95} & \makebox[0pt][c]{\textbf{70.41}} & \makebox[0pt][c]{77.42} & \makebox[0pt][c]{77.10} & \makebox[0pt][c]{78.06} & \makebox[0pt][c]{77.33} & \makebox[0pt][c]{\textbf{79.13}} & \makebox[0pt][c]{82.46} & \makebox[0pt][c]{81.39} & \makebox[0pt][c]{82.14} & \makebox[0pt][c]{81.57} & \makebox[0pt][c]{\textbf{83.78}}\\
    \makebox[0pt][c]{TRADES\(^\text{\cite{ZhaThe19}}\)}  & \makebox[0pt][c]{49.15} & \makebox[0pt][c]{52.62} & \makebox[0pt][c]{49.61} & \makebox[0pt][c]{52.86} & \makebox[0pt][c]{\textbf{53.30}} & \makebox[0pt][c]{59.63} & \makebox[0pt][c]{52.31} & \makebox[0pt][c]{60.45} & \makebox[0pt][c]{62.51} & \makebox[0pt][c]{\textbf{64.05}} & \makebox[0pt][c]{72.06} & \makebox[0pt][c]{75.89} & \makebox[0pt][c]{72.46} & \makebox[0pt][c]{76.10} & \makebox[0pt][c]{\textbf{77.27}} & \makebox[0pt][c]{75.43} & \makebox[0pt][c]{77.26} & \makebox[0pt][c]{76.28} & \makebox[0pt][c]{79.65} & \makebox[0pt][c]{\textbf{81.04}}\\
    \makebox[0pt][c]{MART\(^\text{\cite{Wan20}}\)}  & \makebox[0pt][c]{45.91} & \makebox[0pt][c]{48.16} & \makebox[0pt][c]{46.81} & \makebox[0pt][c]{50.36} & \makebox[0pt][c]{\textbf{52.35}} & \makebox[0pt][c]{59.02} & \makebox[0pt][c]{48.35} & \makebox[0pt][c]{60.36} & \makebox[0pt][c]{63.87} & \makebox[0pt][c]{\textbf{65.89}} & \makebox[0pt][c]{67.47} & \makebox[0pt][c]{72.55} & \makebox[0pt][c]{68.06} & \makebox[0pt][c]{72.09} & \makebox[0pt][c]{\textbf{73.48}} & \makebox[0pt][c]{73.57} & \makebox[0pt][c]{72.63} & \makebox[0pt][c]{74.73} & \makebox[0pt][c]{78.21} & \makebox[0pt][c]{\textbf{79.49}}\\
    \makebox[0pt][c]{HAT\(^\text{\cite{Rad22}}\)}  & \makebox[0pt][c]{46.47} & \makebox[0pt][c]{46.96} & \makebox[0pt][c]{47.04} & \makebox[0pt][c]{47.08} & \makebox[0pt][c]{\textbf{47.94}} & \makebox[0pt][c]{74.95} & \makebox[0pt][c]{74.08} & \makebox[0pt][c]{75.12} & \makebox[0pt][c]{74.22} & \makebox[0pt][c]{\textbf{75.82}} & \makebox[0pt][c]{71.01} & \makebox[0pt][c]{72.27} & \makebox[0pt][c]{71.46} & \makebox[0pt][c]{72.64} & \makebox[0pt][c]{\textbf{74.04}} & \makebox[0pt][c]{83.01} & \makebox[0pt][c]{83.62} & \makebox[0pt][c]{83.97} & \makebox[0pt][c]{83.86} & \makebox[0pt][c]{\textbf{85.22}}\\
    \makebox[0pt][c]{LOAT\(^\text{\cite{Yin24}}\)}  & \makebox[0pt][c]{45.31} & \makebox[0pt][c]{47.67} & \makebox[0pt][c]{46.13} & \makebox[0pt][c]{47.71} & \makebox[0pt][c]{\textbf{48.72}} & \makebox[0pt][c]{73.16} & \makebox[0pt][c]{67.40} & \makebox[0pt][c]{74.90} & \makebox[0pt][c]{73.99} & \makebox[0pt][c]{\textbf{76.35}} & \makebox[0pt][c]{70.11} & \makebox[0pt][c]{72.14} & \makebox[0pt][c]{70.82} & \makebox[0pt][c]{72.28} & \makebox[0pt][c]{\textbf{73.26}} & \makebox[0pt][c]{83.64} & \makebox[0pt][c]{83.12} & \makebox[0pt][c]{84.70} & \makebox[0pt][c]{83.36} & \makebox[0pt][c]{\textbf{85.75}}\\
    \bottomrule
  \end{tabularx}
  }
\end{table*}

\subsection{Attack Performance}
\label{Subsec42}

Attack performance represents the upper bound of an attack method's effectiveness. We compare the attack performance of the proposed SDM and other competing methods against different defenses, with results shown in Tab.~\ref{Tab3}. All attack methods use a total iteration step of 1000. Under the \(\boldsymbol{\ell}_\infty\)-norm constraint, their perturbation budget is 8/255 and the step size is 2/255. Under the \(\boldsymbol{\ell}_2\)-norm, the perturbation budget is 1.0 and the step size is 0.2. Notably, SDM demonstrates the highest attack effectiveness across all attack categories. Overall, its attack performance is higher than PGD, C\&W, APGD\(_1\), and APGD\(_2\) by 3.49\%, 3.77\%, 2.80\%, and 1.57\%, respectively.

\subsection{Attack Cost-Effectiveness}

\begin{figure}[t]
  \centering
  \includegraphics[width=0.92\linewidth]{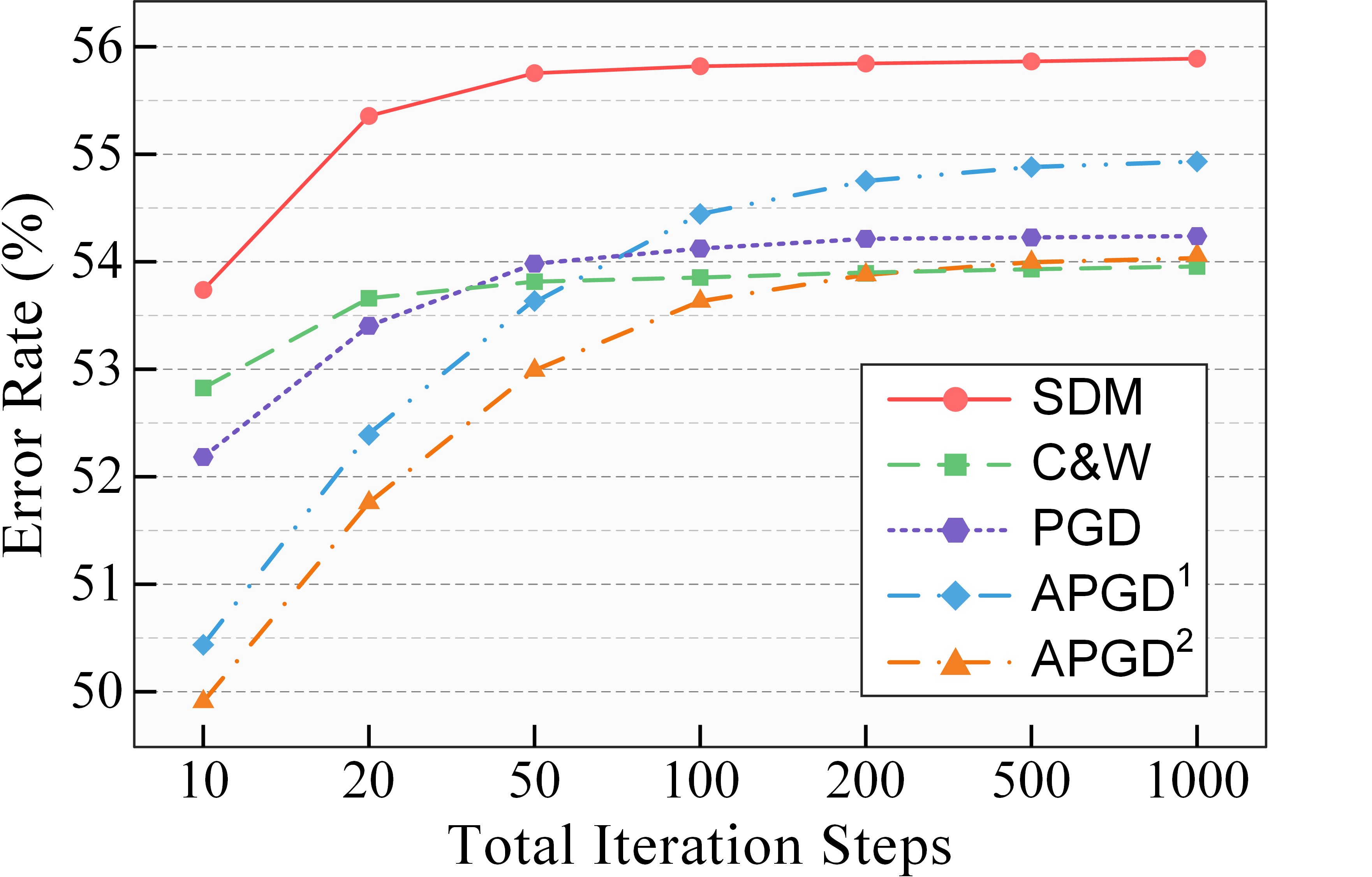}
  \caption{Cost-effectiveness comparison of different attack methods under AT defense on CIFAR-10.}
  \label{Fig2}
\end{figure}

The cost-effectiveness of an attack method represents its practicality to a certain extent. We evaluate the attack effectiveness of SDM and other competing methods under different total iteration steps, with results shown in Fig.~\ref{Fig2}. All attack methods are configured with a perturbation budget of \(\boldsymbol{\ell}_\infty=8/255\) and a step size of \(\boldsymbol{\ell}_\infty=2/255\). It can be found that SDM exhibits significant performance advantages at each total iteration step. This indicates that SDM not only has the strongest upper bound of attack performance but also achieves the highest attack cost-effectiveness.

\subsection{SDM-Based Composite Attacks}

We employ SDM with a total iteration step of 1000 as the fundamental attack method in Auto-PGD (APGD), thereby establishing Auto-SDM (ASDM). For the integrated attack method AA, we replace APGD with ASDM, resulting in AA-SDM. Subsequently, we evaluate these two composite methods under the constraints of \(\boldsymbol{\ell}_\infty=8/255\) and \(\boldsymbol{\ell}_2=1.0\), respectively, with the results presented in Tab.~\ref{Tab4}. It can be observed that after upgrading with SDM, the SDM-based composite attack methods exhibit higher attack performance compared to the original methods. Specifically, ASDM achieves an average attack success rate that is 3.09\% and 2.08\% higher than APGD\(_1\) and APGD\(_2\) in Tab.~\ref{Tab3}, and AA-SDM demonstrates a 1.18\% higher average attack success rate than AA in Tab.~\ref{Tab4}.

\subsection{SDM-Based Defenses}

In prior experiments, the baseline defenses utilized PGD to generate adversarial examples for training.  In this subsection, we employ SDM to assume this role, thereby forming SDM-AT, SDM-TRADES, SDM-MART, SDM-HAT, and SDM-LOAT, respectively. During the training of these defense methods, SDM's perturbation budget, step size, and total iteration step are set to \(\boldsymbol{\ell}_\infty=8/255\), \(\boldsymbol{\ell}_\infty=2/255\), and 10, respectively, consistent with the training configurations in Subsec.~\ref{Subsec41}. Subsequently, we evaluate these SDM-based defenses using multiple attack methods under \(\boldsymbol{\ell}_\infty\)-norm on CIFAR-10, with results summarized in Tab.~\ref{Tab5}. The attack settings are consistent with those in Subsec.~\ref{Subsec42}. Notably, compared to the corresponding data in Tab.~\ref{Tab3}, the error rates of the baseline defenses significantly decreased after SDM enhancement: the average error rates of SDM-based defenses on clean and adversarial examples dropped by 2.03\% and 2.20\%, respectively, compared to PGD-based defenses.

\begin{table}[t]
\vspace{-1pt}
\caption{Error rate comparison (\%) of different defenses under SDM-based composite attack methods on CIFAR-10.}\label{Tab4}
\vspace{-1.2pt}
  \centering
  \newcolumntype{C}{>{\centering}X}
  \newcolumntype{L}[1]{>{\centering\arraybackslash}p{#1}}
  \scalebox{0.87}{
  \begin{tabularx}{1.13\linewidth}{C|L{0.68cm}|*{2}{L{0.7cm}}L{0.9cm}|*{2}{L{0.7cm}}L{0.9cm}}
    \toprule
    \multirow{2}{*}{\makebox[0pt][c]{Defense}} & \multirow{2}{*}{\makebox[0pt][c]{Clean}} & \multicolumn{3}{c|}{\(\boldsymbol{\ell}_\infty=8/255\)} & \multicolumn{3}{c}{\(\boldsymbol{\ell}_2=1.0\)}\\
    & & \makebox[0pt][c]{\scalebox{0.9}[1]{ASDM}} & \makebox[0pt][c]{AA} & \makebox[0pt][c]{\scalebox{0.88}[1]{AA-SDM}} & \makebox[0pt][c]{\scalebox{0.9}[1]{ASDM}} & \makebox[0pt][c]{AA} & \makebox[0pt][c]{\scalebox{0.88}[1]{AA-SDM}}\\
    \midrule
    \makebox[0pt][c]{AT\(^\text{\cite{Mad18}}\)} & \makebox[0pt][c]{\phantom{0}9.75} & \makebox[0pt][c]{56.07} & \makebox[0pt][c]{55.65} & \makebox[0pt][c]{56.28} & \makebox[0pt][c]{71.29} & \makebox[0pt][c]{69.97} & \makebox[0pt][c]{71.86}\\
    \makebox[0pt][c]{TRADES\(^\text{\cite{ZhaThe19}}\)} & \makebox[0pt][c]{17.89} & \makebox[0pt][c]{53.46} & \makebox[0pt][c]{53.02} & \makebox[0pt][c]{53.81} & \makebox[0pt][c]{64.45} & \makebox[0pt][c]{63.30} & \makebox[0pt][c]{64.67}\\
    \makebox[0pt][c]{MART\(^\text{\cite{Wan20}}\)} & \makebox[0pt][c]{18.87} & \makebox[0pt][c]{52.69} & \makebox[0pt][c]{52.16} & \makebox[0pt][c]{52.93} & \makebox[0pt][c]{66.58} & \makebox[0pt][c]{66.22} & \makebox[0pt][c]{67.63}\\
    \makebox[0pt][c]{HAT\(^\text{\cite{Rad22}}\)} & \makebox[0pt][c]{13.27} & \makebox[0pt][c]{48.17} & \makebox[0pt][c]{47.23} & \makebox[0pt][c]{48.37} & \makebox[0pt][c]{76.43} & \makebox[0pt][c]{75.64} & \makebox[0pt][c]{76.94}\\
    \makebox[0pt][c]{LOAT\(^\text{\cite{Yin24}}\)} & \makebox[0pt][c]{15.70} & \makebox[0pt][c]{48.93} & \makebox[0pt][c]{48.41} & \makebox[0pt][c]{49.26} & \makebox[0pt][c]{77.34} & \makebox[0pt][c]{76.09} & \makebox[0pt][c]{77.78}\\
    \bottomrule
  \end{tabularx}
  }
\end{table}

\begin{table}[t]
\caption{Error rate comparison (\%) of SDM-based defenses under various adversarial attack methods on CIFAR-10.}\label{Tab5}
\vspace{-1.2pt}
  \centering
  \newcolumntype{C}{>{\centering}X}
  \newcolumntype{L}[1]{>{\centering\arraybackslash}p{#1}}
  \scalebox{0.87}{
  \begin{tabularx}{1.01\linewidth}{C|L{0.7cm}|*{5}{L{0.7cm}}}
    \toprule
    \makebox[0pt][c]{Defense} & \makebox[0pt][c]{Clean} & \makebox[0pt][c]{PGD} & \makebox[0pt][c]{C\&W} & \makebox[0pt][c]{\scalebox{0.9}[1]{APGD\(_1\)}} & \makebox[0pt][c]{\scalebox{0.9}[1]{APGD\(_2\)}} & \makebox[0pt][c]{SDM} \\
    \midrule
    \makebox[0pt][c]{SDM-AT} & \makebox[0pt][c]{\phantom{0}8.43} & \makebox[0pt][c]{53.44} & \makebox[0pt][c]{53.35} & \makebox[0pt][c]{54.25} & \makebox[0pt][c]{53.96} & \makebox[0pt][c]{54.51}\\
    \makebox[0pt][c]{SDM-TRADES} & \makebox[0pt][c]{16.11} & \makebox[0pt][c]{45.81} & \makebox[0pt][c]{49.47} & \makebox[0pt][c]{45.90} & \makebox[0pt][c]{50.58} & \makebox[0pt][c]{51.12}\\
    \makebox[0pt][c]{SDM-MART} & \makebox[0pt][c]{14.64} & \makebox[0pt][c]{40.51} & \makebox[0pt][c]{47.40} & \makebox[0pt][c]{41.86} & \makebox[0pt][c]{49.76} & \makebox[0pt][c]{50.17}\\
    \makebox[0pt][c]{SDM-HAT} & \makebox[0pt][c]{12.90} & \makebox[0pt][c]{42.23} & \makebox[0pt][c]{46.26} & \makebox[0pt][c]{43.54} & \makebox[0pt][c]{46.51} & \makebox[0pt][c]{47.03}\\
    \makebox[0pt][c]{SDM-LOAT} & \makebox[0pt][c]{13.24} & \makebox[0pt][c]{41.63} & \makebox[0pt][c]{45.24} & \makebox[0pt][c]{43.09} & \makebox[0pt][c]{45.75} & \makebox[0pt][c]{46.85}\\
    \bottomrule
  \end{tabularx}
  }
\end{table}

\section{Conclusion}
\label{Sec5}

In this paper, we propose SDM for efficiently generating adversarial examples in white-box scenarios. Compared with previous adversarial attack methods, SDM employs a more challenging optimization objective. Therefore, we design a novel optimization procedure based on the idea of sequential optimization, which relies on a three-layer optimization framework of "cycle-stage-step". This framework gradually approximates the optimal solution by applying different loss functions at various optimization stages. Experimental results demonstrate that SDM exhibits high effectiveness and practicality, surpassing previous SOTA methods in terms of attack performance and cost-effectiveness. In addition to evaluating model robustness, SDM can be integrated with adversarial training \hspace{-0.3pt}{and} \hspace{-0.3pt}{related} \hspace{-0.3pt}{techniques} \hspace{-0.3pt}{to} \hspace{-0.3pt}{enhance} \hspace{-0.3pt}{defensive} \hspace{-0.3pt}{capabilities.}

\balance

\section*{GenAI Usage Disclosure}

No GenAI tools were used in any stage of the research, nor in the writing.

\normalem
\bibliographystyle{ACM-Reference-Format}
\bibliography{sample-base}


\begin{thebibliography}{20}


\ifx \showCODEN    \undefined \def \showCODEN     #1{\unskip}     \fi
\ifx \showISBNx    \undefined \def \showISBNx     #1{\unskip}     \fi
\ifx \showISBNxiii \undefined \def \showISBNxiii  #1{\unskip}     \fi
\ifx \showISSN     \undefined \def \showISSN      #1{\unskip}     \fi
\ifx \showLCCN     \undefined \def \showLCCN      #1{\unskip}     \fi
\ifx \shownote     \undefined \def \shownote      #1{#1}          \fi
\ifx \showarticletitle \undefined \def \showarticletitle #1{#1}   \fi
\ifx \showURL      \undefined \def \showURL       {\relax}        \fi
\providecommand\bibfield[2]{#2}
\providecommand\bibinfo[2]{#2}
\providecommand\natexlab[1]{#1}
\providecommand\showeprint[2][]{arXiv:#2}

\bibitem[LeCun et~al\mbox{.}(2015)]%
        {LeC15}
\bibfield{author}{\bibinfo{person}{Yann LeCun}, \bibinfo{person}{Yoshua Bengio}, {and} \bibinfo{person}{Geoffrey Hinton}.} \bibinfo{year}{2015}\natexlab{}.
\newblock \showarticletitle{Deep learning}.
\newblock \bibinfo{journal}{\emph{Nature}}  \bibinfo{volume}{521} (\bibinfo{date}{May} \bibinfo{year}{2015}), \bibinfo{pages}{436–444}.
\newblock
\href{https://doi.org/10.1038/nature14539}{doi:\nolinkurl{10.1038/nature14539}}


\bibitem[He et~al\mbox{.}(2016)]%
        {HeDee16}
\bibfield{author}{\bibinfo{person}{Kaiming He}, \bibinfo{person}{Xiangyu Zhang}, \bibinfo{person}{Shaoqing Ren}, {and} \bibinfo{person}{Jian Sun}.} \bibinfo{year}{2016}\natexlab{}.
\newblock \showarticletitle{Deep Residual Learning for Image Recognition}. In \bibinfo{booktitle}{\emph{2016 IEEE/CVF Conference on Computer Vision and Pattern Recognition (CVPR)}}. \bibinfo{pages}{770--778}.
\newblock
\href{https://doi.org/10.1109/CVPR.2016.90}{doi:\nolinkurl{10.1109/CVPR.2016.90}}


\bibitem[Chen et~al\mbox{.}(2021)]%
        {CheRGB21}
\bibfield{author}{\bibinfo{person}{Qian Chen}, \bibinfo{person}{Ze Liu}, \bibinfo{person}{Yi Zhang}, \bibinfo{person}{Keren Fu}, \bibinfo{person}{Qijun Zhao}, {and} \bibinfo{person}{Hongwei Du}.} \bibinfo{year}{2021}\natexlab{}.
\newblock \showarticletitle{{RGB-D} Salient Object Detection via 3D Convolutional Neural Networks}. In \bibinfo{booktitle}{\emph{Thirty-Third Conference on Innovative Applications of Artificial Intelligence, {IAAI} 2021, Virtual Event, February 2-9, 2021}}. \bibinfo{publisher}{{AAAI} Press}, \bibinfo{pages}{1063--1071}.
\newblock
\href{https://doi.org/10.1609/AAAI.V35I2.16191}{doi:\nolinkurl{10.1609/AAAI.V35I2.16191}}


\bibitem[Jin et~al\mbox{.}(2021)]%
        {Jin21}
\bibfield{author}{\bibinfo{person}{Ge Jin}, \bibinfo{person}{Chuancai Liu}, {and} \bibinfo{person}{Xu Chen}.} \bibinfo{year}{2021}\natexlab{}.
\newblock \showarticletitle{Adversarial network integrating dual attention and sparse representation for semi-supervised semantic segmentation}.
\newblock \bibinfo{journal}{\emph{Inf. Process. Manag.}} \bibinfo{volume}{58}, \bibinfo{number}{5} (\bibinfo{year}{2021}), \bibinfo{pages}{102680}.
\newblock
\href{https://doi.org/10.1016/J.IPM.2021.102680}{doi:\nolinkurl{10.1016/J.IPM.2021.102680}}


\bibitem[Neocleous and Schizas(2014)]%
        {Neo14}
\bibfield{author}{\bibinfo{person}{Costas Neocleous} {and} \bibinfo{person}{Christos~N. Schizas}.} \bibinfo{year}{2014}\natexlab{}.
\newblock \showarticletitle{On the Claim for the Existence of "Adversarial Examples" in Deep Learning Neural Networks}. In \bibinfo{booktitle}{\emph{{NCTA} 2014 - Proceedings of the International Conference on Neural Computation Theory and Applications, part of {IJCCI} 2014, Rome, Italy, 22 - 24 October, 2014}}, \bibfield{editor}{\bibinfo{person}{Kurosh Madani} {and} \bibinfo{person}{Joaquim Filipe}} (Eds.). \bibinfo{publisher}{SciTePress}, \bibinfo{pages}{306--309}.
\newblock
\href{https://doi.org/10.5220/0005152503060309}{doi:\nolinkurl{10.5220/0005152503060309}}


\bibitem[Goodfellow et~al\mbox{.}(2015)]%
        {Goo15}
\bibfield{author}{\bibinfo{person}{Ian~J. Goodfellow}, \bibinfo{person}{Jonathon Shlens}, {and} \bibinfo{person}{Christian Szegedy}.} \bibinfo{year}{2015}\natexlab{}.
\newblock \showarticletitle{Explaining and Harnessing Adversarial Examples}. In \bibinfo{booktitle}{\emph{2015 International Conference on Learning Representations (ICLR)}}, \bibfield{editor}{\bibinfo{person}{Yoshua Bengio} {and} \bibinfo{person}{Yann LeCun}} (Eds.).
\newblock


\bibitem[Croce and Hein(2020)]%
        {Cro20}
\bibfield{author}{\bibinfo{person}{Francesco Croce} {and} \bibinfo{person}{Matthias Hein}.} \bibinfo{year}{2020}\natexlab{}.
\newblock \showarticletitle{Reliable evaluation of adversarial robustness with an ensemble of diverse parameter-free attacks}. In \bibinfo{booktitle}{\emph{2020 International Conference on Machine Learning (ICML)}}, Vol.~\bibinfo{volume}{119}. \bibinfo{publisher}{{PMLR}}, \bibinfo{pages}{2206--2216}.
\newblock


\bibitem[Wang et~al\mbox{.}(2024)]%
        {Wan24}
\bibfield{author}{\bibinfo{person}{Xinyi Wang}, \bibinfo{person}{Zhibo Jin}, \bibinfo{person}{Zhiyu Zhu}, \bibinfo{person}{Jiayu Zhang}, {and} \bibinfo{person}{Huaming Chen}.} \bibinfo{year}{2024}\natexlab{}.
\newblock \showarticletitle{Improving Adversarial Transferability via Frequency-Guided Sample Relevance Attack}. In \bibinfo{booktitle}{\emph{Proceedings of the 33rd {ACM} International Conference on Information and Knowledge Management, {CIKM} 2024, Boise, ID, USA, October 21-25, 2024}}. \bibinfo{publisher}{{ACM}}, \bibinfo{pages}{2410--2419}.
\newblock
\href{https://doi.org/10.1145/3627673.3679858}{doi:\nolinkurl{10.1145/3627673.3679858}}


\bibitem[Akhtar et~al\mbox{.}(2021)]%
        {Akh21}
\bibfield{author}{\bibinfo{person}{Naveed Akhtar}, \bibinfo{person}{Ajmal Mian}, \bibinfo{person}{Navid Kardan}, {and} \bibinfo{person}{Mubarak Shah}.} \bibinfo{year}{2021}\natexlab{}.
\newblock \showarticletitle{Threat of Adversarial Attacks on Deep Learning in Computer Vision: Survey {II}}.
\newblock \bibinfo{journal}{\emph{CoRR}}  \bibinfo{volume}{abs/2108.00401} (\bibinfo{year}{2021}).
\newblock
\showeprint[arXiv]{2108.00401}


\bibitem[Madry et~al\mbox{.}(2018)]%
        {Mad18}
\bibfield{author}{\bibinfo{person}{Aleksander Madry}, \bibinfo{person}{Aleksandar Makelov}, \bibinfo{person}{Ludwig Schmidt}, \bibinfo{person}{Dimitris Tsipras}, {and} \bibinfo{person}{Adrian Vladu}.} \bibinfo{year}{2018}\natexlab{}.
\newblock \showarticletitle{Towards Deep Learning Models Resistant to Adversarial Attacks}. In \bibinfo{booktitle}{\emph{2018 International Conference on Learning Representations (ICLR)}}. \bibinfo{publisher}{OpenReview.net}.
\newblock


\bibitem[Szegedy et~al\mbox{.}(2014)]%
        {Chr14}
\bibfield{author}{\bibinfo{person}{Christian Szegedy}, \bibinfo{person}{Wojciech Zaremba}, \bibinfo{person}{Ilya Sutskever}, \bibinfo{person}{Joan Bruna}, \bibinfo{person}{Dumitru Erhan}, \bibinfo{person}{Ian~J. Goodfellow}, {and} \bibinfo{person}{Rob Fergus}.} \bibinfo{year}{2014}\natexlab{}.
\newblock \showarticletitle{Intriguing properties of neural networks}. In \bibinfo{booktitle}{\emph{2nd International Conference on Learning Representations, {ICLR} 2014, Banff, AB, Canada, April 14-16, 2014, Conference Track Proceedings}}.
\newblock
\urldef\tempurl%
\url{http://arxiv.org/abs/1312.6199}
\showURL{%
\tempurl}


\bibitem[Kurakin et~al\mbox{.}(2017)]%
        {Kur17}
\bibfield{author}{\bibinfo{person}{Alexey Kurakin}, \bibinfo{person}{Ian~J. Goodfellow}, {and} \bibinfo{person}{Samy Bengio}.} \bibinfo{year}{2017}\natexlab{}.
\newblock \showarticletitle{Adversarial examples in the physical world}. In \bibinfo{booktitle}{\emph{2017 International Conference on Learning Representations (ICLR)}}. \bibinfo{publisher}{OpenReview.net}.
\newblock


\bibitem[Carlini and Wagner(2017)]%
        {Car17}
\bibfield{author}{\bibinfo{person}{Nicholas Carlini} {and} \bibinfo{person}{David~A. Wagner}.} \bibinfo{year}{2017}\natexlab{}.
\newblock \showarticletitle{Towards Evaluating the Robustness of Neural Networks}. In \bibinfo{booktitle}{\emph{2017 {IEEE} Symposium on Security and Privacy, {SP} 2017, San Jose, CA, USA, May 22-26, 2017}}. \bibinfo{publisher}{{IEEE} Computer Society}, \bibinfo{pages}{39--57}.
\newblock
\href{https://doi.org/10.1109/SP.2017.49}{doi:\nolinkurl{10.1109/SP.2017.49}}


\bibitem[Spall(1992)]%
        {Jam92}
\bibfield{author}{\bibinfo{person}{J.C. Spall}.} \bibinfo{year}{1992}\natexlab{}.
\newblock \showarticletitle{Multivariate stochastic approximation using a simultaneous perturbation gradient approximation}.
\newblock \bibinfo{journal}{\emph{IEEE Trans. Automat. Control}} \bibinfo{volume}{37}, \bibinfo{number}{3} (\bibinfo{year}{1992}), \bibinfo{pages}{332--341}.
\newblock
\href{https://doi.org/10.1109/9.119632}{doi:\nolinkurl{10.1109/9.119632}}


\bibitem[Zagoruyko and Komodakis(2016)]%
        {Zag16}
\bibfield{author}{\bibinfo{person}{Sergey Zagoruyko} {and} \bibinfo{person}{Nikos Komodakis}.} \bibinfo{year}{2016}\natexlab{}.
\newblock \showarticletitle{Wide Residual Networks}. In \bibinfo{booktitle}{\emph{BMVC, York, UK, September 19-22}}.
\newblock
\urldef\tempurl%
\url{https://bmva-archive.org.uk/bmvc/2016/papers/paper087/index.html}
\showURL{%
\tempurl}


\bibitem[Krizhevsky and Hinton(2009)]%
        {Kri09}
\bibfield{author}{\bibinfo{person}{A. Krizhevsky} {and} \bibinfo{person}{G. Hinton}.} \bibinfo{year}{2009}\natexlab{}.
\newblock \showarticletitle{Learning multiple layers of features from tiny images}.
\newblock \bibinfo{journal}{\emph{Master's thesis, Department of Computer Science, University of Toronto}} (\bibinfo{year}{2009}).
\newblock
\urldef\tempurl%
\url{https://www.cs.utoronto.ca/~kriz/cifar.html}
\showURL{%
\tempurl}


\bibitem[Zhang et~al\mbox{.}(2019)]%
        {ZhaThe19}
\bibfield{author}{\bibinfo{person}{Hongyang Zhang}, \bibinfo{person}{Yaodong Yu}, \bibinfo{person}{Jiantao Jiao}, \bibinfo{person}{Eric~P. Xing}, \bibinfo{person}{Laurent~El Ghaoui}, {and} \bibinfo{person}{Michael~I. Jordan}.} \bibinfo{year}{2019}\natexlab{}.
\newblock \showarticletitle{Theoretically Principled Trade-off between Robustness and Accuracy}. In \bibinfo{booktitle}{\emph{Proceedings of the 36th International Conference on Machine Learning, {ICML} 2019, 9-15 June 2019, Long Beach, California, {USA}}} \emph{(\bibinfo{series}{Proceedings of Machine Learning Research}, Vol.~\bibinfo{volume}{97})}, \bibfield{editor}{\bibinfo{person}{Kamalika Chaudhuri} {and} \bibinfo{person}{Ruslan Salakhutdinov}} (Eds.). \bibinfo{publisher}{{PMLR}}, \bibinfo{pages}{7472--7482}.
\newblock
\urldef\tempurl%
\url{http://proceedings.mlr.press/v97/zhang19p.html}
\showURL{%
\tempurl}


\bibitem[Wang et~al\mbox{.}(2020)]%
        {Wan20}
\bibfield{author}{\bibinfo{person}{Yisen Wang}, \bibinfo{person}{Difan Zou}, \bibinfo{person}{Jinfeng Yi}, \bibinfo{person}{James Bailey}, \bibinfo{person}{Xingjun Ma}, {and} \bibinfo{person}{Quanquan Gu}.} \bibinfo{year}{2020}\natexlab{}.
\newblock \showarticletitle{Improving Adversarial Robustness Requires Revisiting Misclassified Examples}. In \bibinfo{booktitle}{\emph{8th International Conference on Learning Representations, {ICLR} 2020, Addis Ababa, April 26-30, 2020}}. \bibinfo{publisher}{OpenReview.net}.
\newblock
\urldef\tempurl%
\url{https://openreview.net/forum?id=rklOg6EFwS}
\showURL{%
\tempurl}


\bibitem[Rade and Moosavi{-}Dezfooli(2022)]%
        {Rad22}
\bibfield{author}{\bibinfo{person}{Rahul Rade} {and} \bibinfo{person}{Seyed{-}Mohsen Moosavi{-}Dezfooli}.} \bibinfo{year}{2022}\natexlab{}.
\newblock \showarticletitle{Reducing Excessive Margin to Achieve a Better Accuracy vs. Robustness Trade-off}. In \bibinfo{booktitle}{\emph{The Tenth International Conference on Learning Representations, {ICLR} 2022, Virtual Event, April 25-29, 2022}}. \bibinfo{publisher}{OpenReview.net}.
\newblock
\urldef\tempurl%
\url{https://openreview.net/forum?id=Azh9QBQ4tR7}
\showURL{%
\tempurl}


\bibitem[Yin and Ruan(2024)]%
        {Yin24}
\bibfield{author}{\bibinfo{person}{Xiangyu Yin} {and} \bibinfo{person}{Wenjie Ruan}.} \bibinfo{year}{2024}\natexlab{}.
\newblock \showarticletitle{Boosting Adversarial Training via Fisher-Rao Norm-Based Regularization}. In \bibinfo{booktitle}{\emph{{IEEE/CVF} Conference on Computer Vision and Pattern Recognition, {CVPR} 2024, Seattle, WA, USA, June 16-22, 2024}}. \bibinfo{publisher}{{IEEE}}, \bibinfo{pages}{24544--24553}.
\newblock
\href{https://doi.org/10.1109/CVPR52733.2024.02317}{doi:\nolinkurl{10.1109/CVPR52733.2024.02317}}


\end{thebibliography}

\end{document}